\documentclass{article}
\usepackage{spconf,amsmath,graphicx}

\usepackage{makecell}
\usepackage{diagbox}
\usepackage{multicol}
\usepackage[noend]{algpseudocode}
\usepackage{algorithmicx,algorithm}
\usepackage{bm}

\title{MULTI-LEVEL CONTRASTIVE LEARNING FOR CROSS-LINGUAL ALIGNMENT}
\name{Beiduo Chen\textsuperscript{1}, Wu Guo\textsuperscript{1}, Bin Gu\textsuperscript{1}, Quan Liu\textsuperscript{2}, Yongchao Wang\textsuperscript{2}\thanks{Thanks to the National Natural Science Foundation of China (Grant No. U1836219) for funding.}}
\address{\textsuperscript{1}NELSLIP, University of Science and Technology of China, Hefei, China \\
\textsuperscript{2}State Key Laboratory of Cognitive Intelligence, iFLYTEK Research
}

\begin{document}
%
\maketitle
\begin{abstract}
Cross-language pre-trained models such as multilingual BERT (mBERT) have achieved significant performance in various cross-lingual downstream NLP tasks. This paper proposes a multi-level contrastive learning (ML-CTL) framework to further improve the cross-lingual ability of pre-trained models. The proposed method uses translated parallel data to encourage the model to generate similar semantic embeddings for different languages. However, unlike the sentence-level alignment used in most previous studies, in this paper, we explicitly integrate the word-level information of each pair of parallel sentences into contrastive learning. Moreover, cross-zero noise contrastive estimation (CZ-NCE) loss is proposed to alleviate the impact of the floating-point error in the training process with a small batch size. The proposed method significantly improves the cross-lingual transfer ability of our basic model (mBERT) and outperforms on multiple zero-shot cross-lingual downstream tasks compared to the same-size models in the Xtreme benchmark.
\end{abstract}
\vspace{-5mm}
\section{Introduction}
\label{sec:intro}
\vspace{-3mm}
Recently, cross-lingual pre-trained language models with the structure of transformers like multilingual BERT (mBERT) and cross-lingual language model (XLM) have enabled effective cross-lingual transfer and performed surprisingly well on plenty of downstream tasks \cite{devlin2018bert,lample2019cross}. These models are firstly pre-trained on large-scale corpus covers over 100 languages mainly using the multilingual masked language modeling (MMLM) \cite{devlin2018bert} algorithm, and then fine-tuned on English-supervised training data for downstream tasks aiming to improve the performance on low-resource languages \cite{ruder2019survey}. However, such a pre-training method only encourages implicit cross-lingual alignment in the vector space \cite{muller2021first,de2020s}. Although XLM \cite{lample2019cross} introduced the translation language modeling (TLM) method to further enhance cross-lingual alignment, some scholars believe that TLM only learns the structural pattern of parallel sentence pairs without truly understanding the meaning of sentences \cite{wei2020learning,conneau2019unsupervised}.

A feasible approach to improve the model’s ability on cross-lingual transfer is to incorporate translated parallel sentences into pre-training for explicit alignment. Many studies have used the contrastive learning (CTL) \cite{le2020contrastive,mnih2012fast} to solve this problem as translated parallel sentences can naturally be used as positive pairs in CTL for encouraging the model to learn explicit cross-lingual alignment \cite{chi2020infoxlm,wang2021aligning}. However, most of them directly use the [CLS] token in BERT \cite{devlin2018bert} to represent the meanings of sentences for cross-language alignment only at the sentence-level, without considering alignment at the word-level. Wei \emph{et al.} use the idea of bag-of-words to address this problem \cite{wei2020learning}, however, where the advantage of BERT’s bidirectional structure to generate dynamic contextual word embeddings is overshadowed by the fixed word embeddings.

To overcome these problems, we propose a multi-level contrastive learning (ML-CTL) method to integrate both sentence-level and word-level cross-lingual alignment into one training framework. In concrete terms, we first construct samples that consist both of sentences and positions of words, and then fed them into a designed encoder that contains a basic language model to get concatenated contextual embeddings (CCE) representing both sentence information and contextual word information. After pre-trained with CCEs by contrastive learning, the basic model’s cross-lingual ability can be significantly improved by learning the structural pattern of parallel sentence pairs as well as aligning the semantic meanings of parallel words in the sentences.

In practice, CTL with commonly used information noise contrastive estimation (infoNCE) \cite{oord2018representation} loss requires a large batch size to improve performance \cite{chen2020simple} which is too expensive for cases with limited computational resources. However, the infoNCE loss value of a mini-batch will soon approach zero during training with a small batch size and the floating-point error will seriously deteriorate the final performance \cite{chen2021simpler}. Hence, we propose cross-zero NCE (CZ-NCE) loss by modifying the lower bound of the infoNCE loss to alleviate the impact of the floating-point error.

We carry out evaluation experiments for the proposed pre-trained models on multiple zero-shot cross-lingual tasks in the Xtreme benchmark \cite{hu2020xtreme}. The results demonstrate that the proposed methods achieve a significant performance improvement on a strong baseline.

\begin{figure*}[htb]

\begin{minipage}[b]{.48\linewidth}
  \centering
  \centerline{\includegraphics[width=8.5cm]{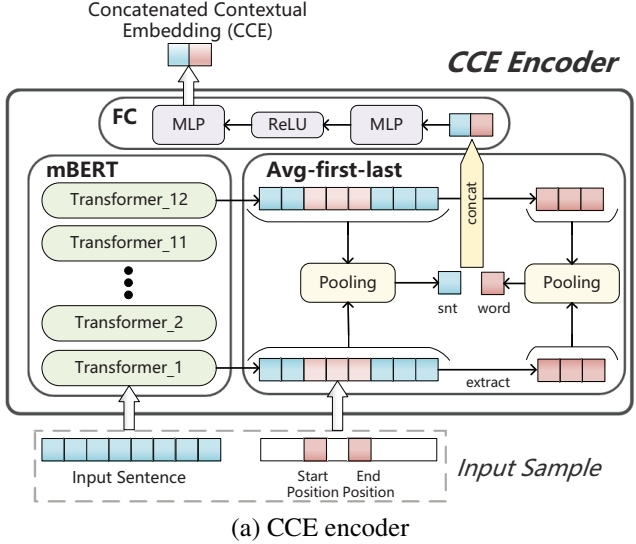}}
  \centerline{(a) CCE encoder}\medskip
\end{minipage}
\hfill
\begin{minipage}[b]{0.48\linewidth}
  \centering
  \centerline{\includegraphics[width=8.5cm]{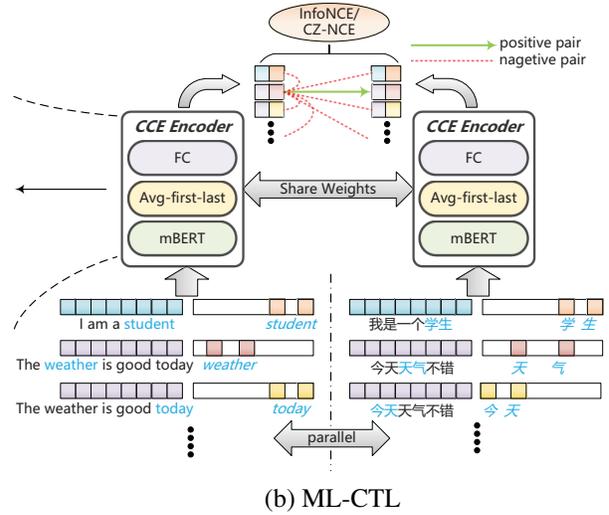}}
  \centerline{(b) ML-CTL}\medskip
\end{minipage}
\vspace{-5mm}
\caption{(a) shows the internal structure of the CCE encoder. We input a sentence with a word’s precise position information into the encoder to obtain a concatenated embedding. (b) is the concept map of ML-CTL which illustrates how to pre-train the model on both sentence-level and word-level with contrastive learning.}\label{Fig1}
\vspace{-4mm}
\end{figure*}

In summary, we present two novel contributions:

\noindent (1) ML-CTL is proposed to improve the cross-lingual alignment ability of pre-trained language models by applying contrastive learning on concatenated contextual embeddings which contain information of both sentences and words.

\noindent (2) CZ-NCE is proposed to alleviate the impact of the floating-point error with a small training batch size.

\vspace{-3mm}
\section{MULTI-LEVEL CONTRASTIVE LEARNING}
\label{sec:method1}
\vspace{-3mm}
To integrate the word-level alignment information into the sentence-level cross-lingual pre-training, the word positioned sample (WPS) is first constructed and then the concatenated contextual embedding (CCE) of WPS is extracted for the following multi-level contrastive learning (ML-CTL).

\vspace{-4mm}
\subsection{Word Positioned Sample Construction}
\label{ssec:WPS}
\vspace{-2mm}
In conventional CTL training, one sentence is considered as a training sample \cite{chi2020infoxlm,wang2021aligning}. For the proposed ML-CTL, the start and end positions of each word (without stop words) are appended to the sentence to form a new word positioned sample (WPS). Using a bilingual dictionary that has been removed stop words, a parallel WPSs pair is obtained with accurate word alignment, as Algorithm \ref{Algo1} shows.

\setlength{\textfloatsep}{0.1cm}
\begin{algorithm}[htb]
\caption{Parallel WPSs construction}\label{Algo1} 
\hspace*{0.02in}  
\# use a parallel sentences pair for example\\
Fetch parallel sentences pair \(\{\mathbf{A} \leftrightarrow \mathbf{B}\}\) with words string \(\mathbf{A}=\{a_1,a_2,...,a_N\}, \mathbf{B}=\{b_1,b_2,...,b_M\}\) \\
Bilingual Dictionary \(\mathbf{V}_{\mathbf{A}\times\mathbf{B}} \{a_i \leftrightarrow b_i \}  \) \\
TK = BertTokenizer\cite{sennrich2015neural,devlin2018bert}  \# a tokenizer for BERT
\begin{algorithmic}[1]
\State \# get parallel words pair and their positions 
\For{\(a_i \) in \( \mathbf{V}_\mathbf{A} \)} 
\If{\(\mathbf{A}\).count(' \(a_i\) ')==TK(\(\mathbf{A}\)).count(TK('\(a_i\)'))==1} 
\State \# ‘count’ returns the occurrence number
\State get \(\mathbf{S}=(\mathbf{A},{start}_{a_i},{end}_{a_i}) \)
\State \# record the word’s position in the sentence
\State get \(b_i=\mathbf{V}_{\mathbf{A}\times\mathbf{B}}[a_i]\)
\If{\(\mathbf{B}\).count(' \(b_i\) ')==1}
\If{TK(\(\mathbf{B}\)).count(TK('\(b_i\)'))==1}
\State get \(\mathbf{S}^\mathbf{t}=(\mathbf{B},{start}_{b_i},{end}_{b_i}) \)
\State \Return \(\mathbf{S},\mathbf{S}^\mathbf{t}\) 
\State \# return a pair of positive parallel WPSs
\EndIf
\EndIf
\EndIf
\EndFor
\end{algorithmic}
\end{algorithm}
\setlength{\floatsep}{0.1cm}

\vspace{-4mm}
\subsection{Concatenated Contextual Embedding}
\label{ssec:CCE}
\vspace{-2mm}
As shown in Fig. \ref{Fig1}a, mBERT as part of the encoder is applied to extract concatenated contextual embedding (CCE). For the sentence-level representation, we choose the average of all tokens from the first and last transformer layers of mBERT (\emph{avg-first-last}), as it has been proved to be a good way to represent the meaning of a sentence \cite{li2020sentence}. Because high-frequency words may dominate the meaning of the whole sentence \cite{yan2021consert}, it is not a good idea to rely only on sentence embeddings for cross-lingual alignment.

Using WPS, we can extract the word embedding that has contextual semantics due to the BERT’s bidirectional structure. The extracted word embedding is more suitable to be the representation for alignment than the conventional fixed word embedding. To be consistent with the sentence embedding, we also adapt \emph{avg-first-last} pattern.

Finally, these two embeddings are spliced and fed into the following FC layer to obtain the final CCE.

\vspace{-4mm}
\subsection{Multi-Level Contrastive Learning}
\label{ssec:MLCTL}
\vspace{-2mm}
As shown in Fig. \ref{Fig1}b, for each batch of parallel WPSs \((\mathbf{X},\mathbf{Y})\) in two languages, we input them separately to the encoder to obtain CCEs \(\mathbf{X}_{c}=\{\mathbf{x}_{{c1}},\mathbf{x}_{{c}{2}},...,\mathbf{x}_{{cn}}\}, \mathbf{Y}_{c}=\{\mathbf{y}_{{c}{1}},\mathbf{y}_{{c}{2}},...,\mathbf{y}_{{cn}}\}\) where \({n}\) is the batch size. Each \(\mathbf{y}_{{ci}}\) is treated as a positive sample \(\mathbf{k}^+\) for \(\mathbf{x}_{{ci}}\) while a batch of all others \( \{{{\mathbf{X}}_{c}}^{/\mathbf{x}_{{ci}}}\cup{\mathbf{Y}_{c}}^{/\mathbf{y}_{{ci}}}\}\) are considered as negative samples \(\{\mathbf{k}^-\}\) ( \({\mathbf{X}_{c}}^{/\mathbf{x}_{{ci}}}\) denotes the remaining instances of \(\mathbf{X}_{c}\) without \(\mathbf{x}_{{ci}}\)). Utilizing infoNCE loss \cite{oord2018representation}, the optimization target for each \(\mathbf{x}_{{ci}}\) is achieved:
\begin{equation}\label{eq1}
\setlength{\abovedisplayskip}{0pt}
\setlength{\belowdisplayskip}{0pt}
{L}_{{info}}(\mathbf{x}_{{ci}})=-log(\frac{e^{s(\mathbf{x}_{{ci}},\mathbf{y}_{{ci}}\ )}}{e^{s(\mathbf{x}_{{ci}},\mathbf{y}_{{ci}})}+\sum_{{\mathbf{k}_{j}}^-}^{\{\mathbf{k}^-\}}e^{s(\mathbf{x}_{{ci}},{\mathbf{k}_{j}}^-)}})
\end{equation}
\noindent where  \(s(\cdot)\) denotes the cosine similarity calculation with a denominator (temperature parameter \(t\) set as 0.07).

Then the total loss for a batch of samples is shown as:
\begin{equation}\label{eq2}
\setlength{\abovedisplayskip}{2pt}
\setlength{\belowdisplayskip}{0pt}
{L}_{info\_batch}=\ \frac{\sum_{i}^{n}{{({L}}_{{info}}(\mathbf{x}_{{ci}})+{L}_{{info}}(\mathbf{y}_{{ci}}))}}{2n}
\end{equation}

InfoNCE loss brings positive samples closer together and negative samples farther away, which fits well with cross-lingual alignment. It is worth noting that the WPSs with parallel sentences but different words are also treated as negative pairs, as depicted in Fig. \ref{Fig1}b. We hypothesize that this design is more effective in word-level alignment as the distance between negative word pairs can be further extended. In order not to damage the language information within mBERT itself, we also add MLM \cite{devlin2018bert} as an auxiliary task in the real training. The total loss is calculated as follow:
\begin{equation}\label{eq3}
\setlength{\abovedisplayskip}{2pt}
\setlength{\belowdisplayskip}{2pt}
{L}_{multi_1}={L}_{info\_batch}+{\alpha}\ast{L}_{MLM}
\end{equation}
\noindent where \({\alpha}\) is the proportion of MLM.

\vspace{-7mm}
\section{CROSS-ZERO NCE}
\label{sec:method2}
\vspace{-4mm}
As an accepted conclusion, contrastive learning requires a large batch size to improve the learning effect \cite{chen2020simple}. However, in most cases with limited computational resources, we can only pre-train the model with a small batch size. The model can quickly distinguish positive and negative samples correctly during pre-training with a small batch size because there are too few negative ones as interferences and CCE can further increase the discriminability. Assuming \({s}^+\) and \({s}_{i}^-\) denote distances between positive and negative samples respectively in infoNCE loss function, while \(e^{{s}^+}\) becomes much larger than \(e^{{s}^-}\) during training with a small batch size, the loss will soon approach 0, which is at the same magnitude with the floating-point error. It may seriously affect the training of contrastive learning \cite{chen2021simpler}.
\begin{equation}\label{eq4}
\setlength{\abovedisplayskip}{-1pt}
\setlength{\belowdisplayskip}{-3pt}
{L}_{{infoNCE}}=-log(\frac{e^{{s}^+}}{e^{{s}^+}+\sum e^{{s}_{i}^-}})
\end{equation}

In Eq. \ref{eq4}, the \(e^{{s}^+}\) in the denominator sets a lower bound 0 for infoNCE loss. Therefore, we consider removing the \(e^{{s}^+}\) in the denominator and modifying infoNCE loss to cross-zero NCE (CZ-NCE) loss for keeping the loss value away from zero during most of the training time.
\begin{equation}\label{eq5}
\setlength{\abovedisplayskip}{0pt}
\setlength{\belowdisplayskip}{0pt}
{L}_{CZ-NCE}=-log(\frac{e^{{s}^+}}{\sum e^{{s}_{i}^-}})
\end{equation}

As Eq. \ref{eq5} shows, the learning goal of CZ-NCE is still the same as infoNCE. In the following, we prove the effectiveness of CZ-NCE on alleviating the disturbance of the float-point error. Assuming \(\varphi=\sum e^{{s}_{i}^--{s}^+}\), we perform the following calculation:
\begin{equation}\label{eq6}
\setlength{\abovedisplayskip}{0pt}
\setlength{\belowdisplayskip}{0pt}
{L}_{CZ-NCE}=-log(\frac{e^{{s}^+}}{\sum e^{{s}_{i}^-}})=log(\sum e^{{s}_{i}^--{s}^+})=log(\varphi)
\end{equation}
\begin{equation}\label{eq7}
\setlength{\abovedisplayskip}{-1pt}
\setlength{\belowdisplayskip}{-5pt}
\nabla_\theta{L}_{CZ-NCE}=\nabla_\theta log(\varphi)=\frac{\nabla_\theta\varphi}{\varphi}=\nabla_\theta(\frac{\varphi}{sg(\varphi)})
\end{equation}

\vspace{-2mm}
\noindent where \(\nabla_\theta\) denotes the gradient calculation and \(sg(\cdot)\) stands for the stop-gradient operator that is defined as identity at forward computation time and has zero partial derivatives, thus effectively constraining its operand to a non-updated constant.

Assuming a new loss \(\rho=\frac{\varphi}{sg(\varphi)}\), it is easy to notice that \(\rho\equiv1\) for any batch of inputs (note the gradient of \(\rho\) is not flat) which is far from 0 and less affected by the floating-point error. Since the back-propagation of neural network training only matters with the gradient of loss instead of the value, CZ-NCE has just the same effect with \(\rho\) for model’s training, indicating that CZ-NCE can indeed alleviate the impact of the float-point error.

Same as before, we set the total learning loss as follow:
\begin{equation}\label{eq8}
\setlength{\abovedisplayskip}{0pt}
\setlength{\belowdisplayskip}{-2pt}
{L}_{multi2}={L}_{CZ-NCE\_batch}+{\alpha}\ast{L}_{MLM}
\end{equation}

\vspace{-3mm}
\section{EXPERIMENTS AND RESULTS}
\label{sec:exp}
\vspace{-2mm}
\begin{table*}
\caption{All results of evaluation experiments on pre-trained models.}\label{tab1}
\begin{center}
\begin{tabular}{l|c|c|c|c|c|c}
\Xhline{1.5pt}
Task     & XNLI  & PAWS-X  & POS  & NER  & BUCC &TATOEBA       \\ \hline
Model$\backslash$Metrics    & Acc. (\%)  & Acc. (\%)  & F1 (\%)  & F1 (\%)  & F1 (\%)  & Acc. (\%)  \\ \Xhline{1pt}
\multicolumn{7}{l}{\emph{Main results compared to strong baselines}}          \\ \Xhline{1pt}
mBERT \cite{devlin2018bert} (base)   & 65.4 & 81.9 & 70.3 & 62.2 & 56.7 & 38.7          \\ \hline
XLM \cite{lample2019cross}  & \textbf{69.1} & 80.9 & 70.1 & 61.2 & 56.8 & 32.6 \\ \hline
MMTE \cite{arivazhagan2019massively}  & 67.4 & 81.3 & 72.3 & 58.3 & 59.8 & 37.9  \\ \hline
ML-CTL-CZ (ours)  & 67.8 & \textbf{85.3} & \textbf{72.3} & \textbf{62.9} &\textbf{78.4} & \textbf{43.4}  \\ \Xhline{1pt}
\multicolumn{7}{l}{\emph{Results of ablation study}}           \\ \Xhline{1pt}
mBERT (base)  & 65.4 & 81.9 & 70.3 & 62.2 & 56.7 & 38.7          \\ \hline
\emph{info-snt}  & 66.255 & 84.092 & 71.544 & 62.157 & 76.426 & 41.148   \\ \hline
\emph{CZ-snt}  & 66.862 & 84.485 & 71.733 & 62.337 & 77.403 & 41.751  \\ \hline
ML-CTL-CZ  & \textbf{67.750} & \textbf{85.321} & \textbf{72.289} & \textbf{62.865} & \textbf{78.440} & \textbf{43.389}  \\ \Xhline{1.5pt}
\end{tabular}
\end{center}
\end{table*}

\begin{figure*}[htb]
\vspace{-7.5mm}
\begin{minipage}[b]{0.24\linewidth}
  \centering
  \centerline{\includegraphics[width=4.0cm]{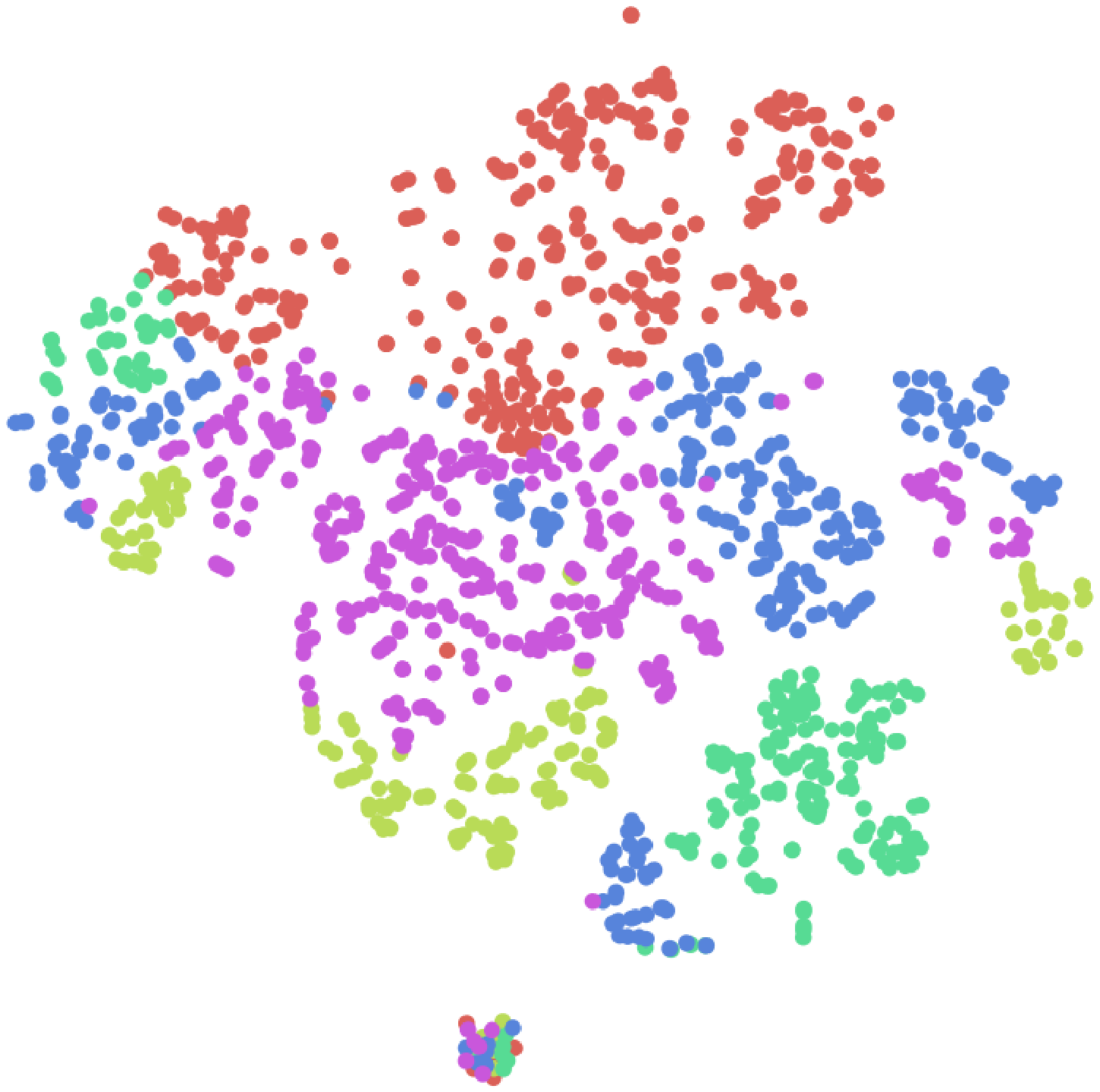}}
  \centerline{(a) mBERT}\medskip
\end{minipage}
\hfill
\begin{minipage}[b]{0.24\linewidth}
  \centering
  \centerline{\includegraphics[width=4.0cm]{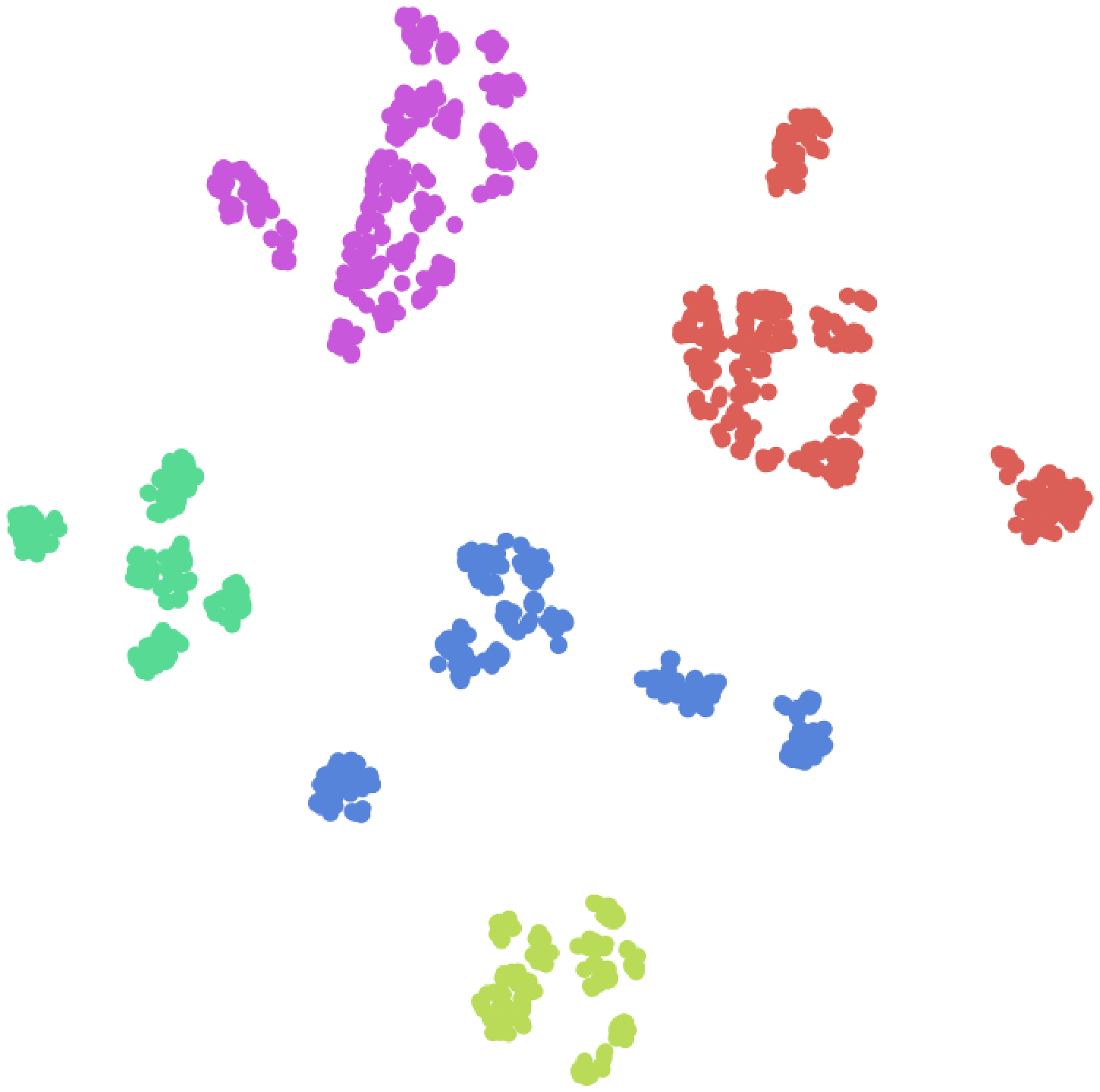}}
  \centerline{(b) info-snt}\medskip
\end{minipage}
\hfill
\begin{minipage}[b]{0.24\linewidth}
  \centering
  \centerline{\includegraphics[width=4.0cm]{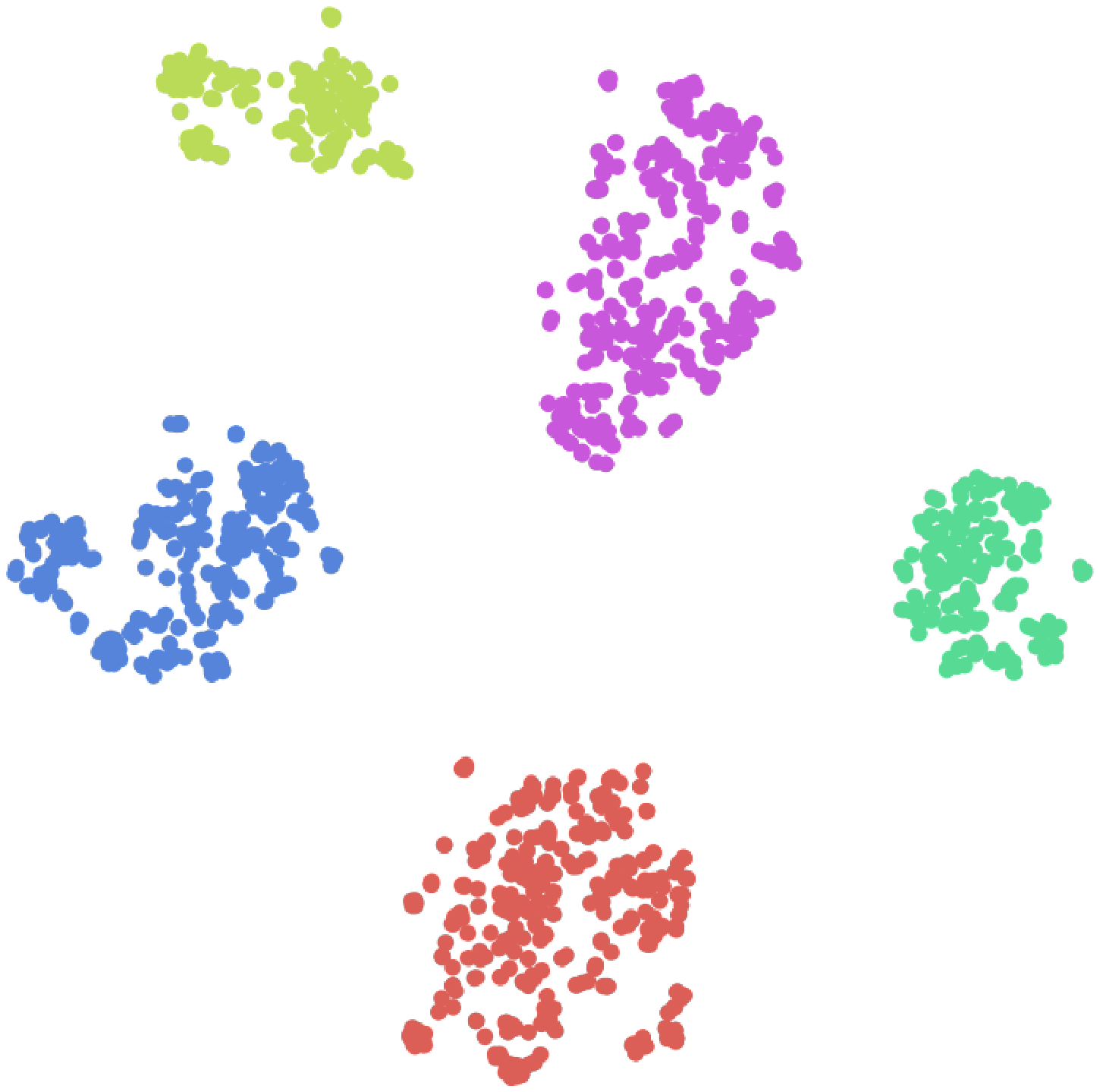}}
  \centerline{(c) CZ-snt}\medskip
\end{minipage}
\hfill
\begin{minipage}[b]{0.24\linewidth}
  \centering
  \centerline{\includegraphics[width=4.0cm]{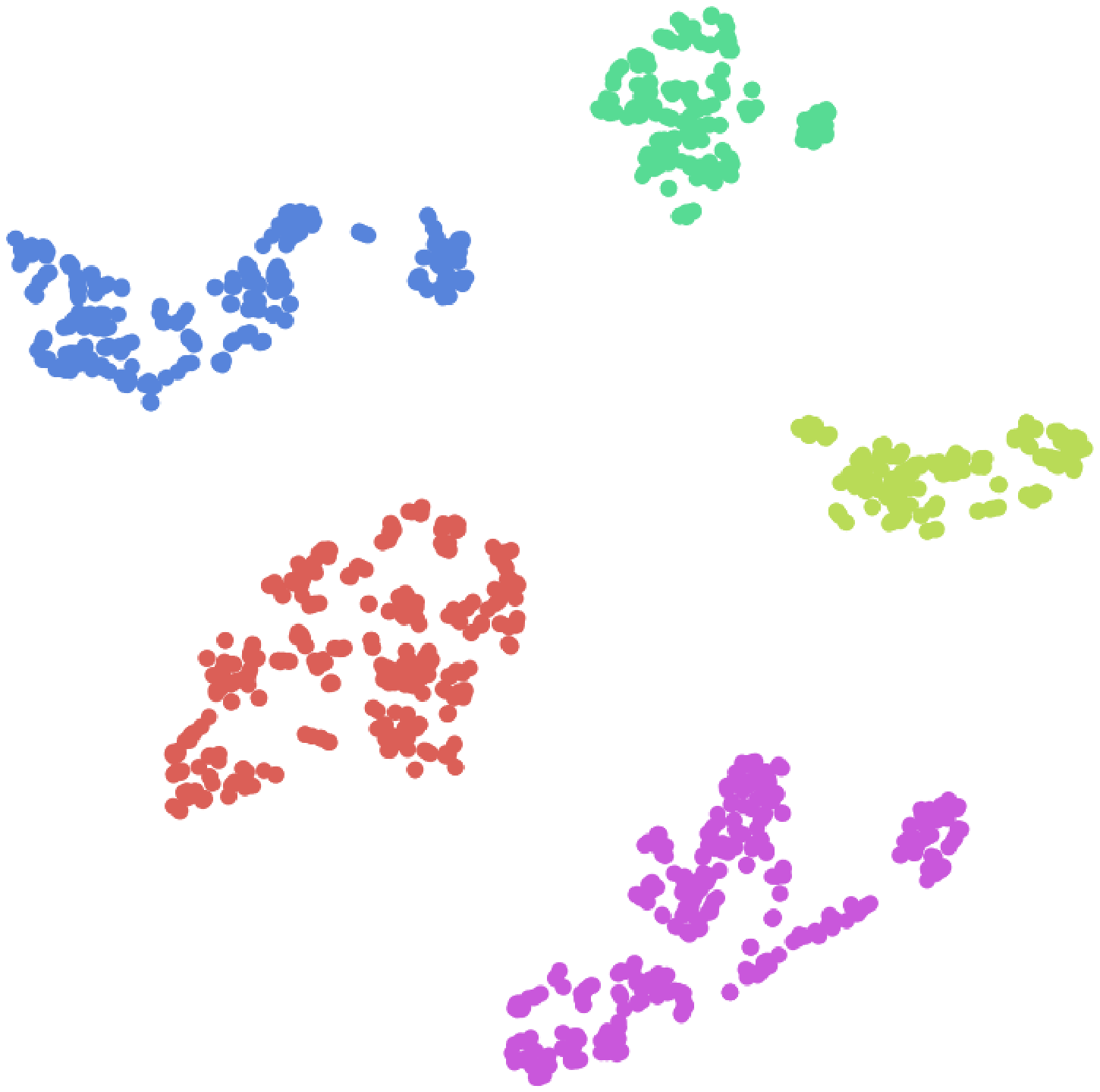}}
  \centerline{(d) ML-CTL-CZ}\medskip
\end{minipage}
\vspace{-5mm}
\caption{Graphs of t-SNE visualization. Each point represents a token embedding and the tokens in sentences across languages with the same meaning have the same color. From the graphs, ML-CTL-CZ has the optimal cross-lingual ability as the distribution of its tokens has better intra-class compactness and inter-class separability.}
\label{Fig2}
\vspace{-4mm}
\end{figure*}

\subsection{Datasets and Experiments Setup}
\label{ssec:data}
\vspace{-2mm}
The United Nations (UN) Parallel Corpus v1.0 is used as a training set, consisting of manually translated UN documents from 1990 to 2014 for the six official UN languages: Arabic, Chinese, English, French, Russian, and Spanish \cite{ziemski2016united}. This dataset has been used for lots of pre-training of LM \cite{conneau2019unsupervised,lample2019cross,wei2020learning}. We create several bilingual dictionaries from Google translate and then randomly construct 250M parallel WPSs in total evenly distributed over six languages. We apply the proposed methods to pre-training the basic model mBERT and generally set the learning rate as 2e-6 while the batch size is 64. We use Adam optimizer \cite{kingma2014adam} and set \({\alpha}\) as 0.1.

To analyze the cross-lingual performance of the proposed pre-trained models, we choose several zero-shot downstream tasks in Xtreme benchmark \cite{hu2020xtreme}: XNLI and PAWS-X for sentence classification with metric as accuracy (Acc.), UD-POS and PANX(NER) for structured prediction with metric as F1 Score(F1), as well as BUCC and Tatoeba for retrieval with metrics as F1 and Acc. All these tasks are widely used in the cross-lingual assessment. We first fine-tune the part of modified mBERT in our pre-trained model on English labeled data and then apply it to predicting on non-English unlabeled data. All the configurations of fine-tuning are set the same as Xtreme benchmark for fair reference.

\vspace{-4mm}
\subsection{Main Results}
\label{ssec:mainresults}
\vspace{-3mm}
For the proposed ML-CTL system with CZ-NCE loss (denoted as ML-CTL-CZ), the CCE encoder is first initialized with mBERT’s parameters and then pre-trained by our methods. We also employ two strong baselines XLM \cite{lample2019cross} and MMTE \cite{arivazhagan2019massively} with the same size as mBERT for comparison. As shown in the upper half of Table \ref{tab1}, the proposed ML-CTL-CZ significantly improves the performance of the basic model (mBERT) and achieves optimal results on multiple downstream tasks compared to the same-size models. The reason for the impressive improvement on BUCC might be that the goal of this task is to look for parallel pairs in a pile of sentences, exactly in line with our pre-training goal.

\vspace{-4mm}
\subsection{Ablation Study}
\label{ssec:ablation}
\vspace{-2mm}
To investigate the effect of the proposed methods, we perform the ablation study as shown in the bottom half of Table \ref{tab1}. All the enhanced systems are initialized with mBERT’s parameters. \emph{info-snt} denotes the pre-trained model that applies only to sentence embeddings (no word embeddings concatenated) with infoNCE loss. \emph{CZ-snt} replaces the former system’s loss with CZ-NCE loss. For the ML-CTL-CZ system, both the word-level and sentence-level information are used with CZ-NCE loss. From the results, CZ- NCE loss can provide improvements over the infoNCE loss in a small training batch size. CCE with multi-level information can increase the discriminability between positive and negative pairs and further improve the cross-language performance of the model.

\vspace{-3mm}
\subsection{T-SNE Visualization for Cross-lingual Ability}
\label{ssec:tSNE}
\vspace{-2mm}
We use t-SNE visualization \cite{van2008visualizing} to intuitively demonstrate the cross-lingual ability of different models in the ablation study. Sentence groups with five different meanings in 6 languages (30 sentences totally) are randomly selected from the UN corpus and fed into different models. Fig. \ref{Fig2} shows the results. Compared with large overlaps among different classes for the basic model mBERT(Fig. \ref{Fig2}a), we observe a clear disentanglement in the feature space with distinctive boundaries between each class for CTL systems. As illustrated, the tokens in Fig. \ref{Fig2}c\&d have better intra-class compactness and inter-class separability than those in Fig. \ref{Fig2}b. The tokens in the same class shown in Fig. \ref{Fig2}d are further closer which indicates that the ML-CTL-CZ has achieved better word-level alignment. These graphs demonstrate the power of the proposed methods.

\vspace{-3mm}
\section{CONCLUSION}
\label{sec:conc}
\vspace{-3mm}
In this paper, we have proposed a framework of multi-level contrastive learning to further help cross-lingual models learn universal representations across languages. As demonstrated by the experiments, word-level information is a strong complement to sentence-level information in cross-lingual alignment. Furthermore, CZ-NCE loss is proposed to reduce the floating-point error near the zero point in the case of training with a small batch size. The proposed model achieves better performance on multiple zero-shot cross-language tasks than the same-size models in Xtreme benchmark. Through the ablation study and t-SNE visualization, we have clearly demonstrated the effectiveness of the proposed methods.

\vfill\pagebreak

\clearpage

\end{document}